\documentclass[letterpaper, 10 pt, conference]{ieeeconf}  %

\IEEEoverridecommandlockouts                              %

\title{\LARGE \bf
Towards Automatic Identification of Globally Valid \\ Geometric Flat Outputs via Numerical Optimization
}

\author{Jake Welde and Vijay Kumar%
\thanks{
	J. Welde and V. Kumar are with the GRASP Laboratory at the University of Pennsylvania. We sincerely thank Dr. Matthew Kvalheim, Dr. Ross Hatton, Dr. Jean Gallier, Dr. Patrick Wensing, and Dr. Justin Carpentier for helpful discussions and suggestions relating to this work.
	We gratefully acknowledge the support of Qualcomm Research, NSF Grant CCR-2112665, and the NSF Graduate Research Fellowship Program.
}
}

\usepackage{cite}

\usepackage{tikzscale}

\usepackage{amsmath,amssymb,tikz-cd}
\usepackage{tikz}

\tikzcdset{every label/.append style = {font = \normalsize}, every matrix/.append style={nodes={font=\normalsize}}}

\newcommand{\metric}[2]{\langle\langle{\hspace{1pt}}#1{\hspace{1pt},\hspace{1pt}}#2{\hspace{1pt}}\rangle\rangle}
\newcommand{\pairing}[2]{\big\langle{\hspace{1pt}}#1{\hspace{1.5pt};\hspace{1.5pt}}#2{\hspace{1pt}}\big\rangle}

\DeclareMathOperator{\proj}{proj}

\DeclareMathOperator{\spn}{span}

\DeclareMathOperator{\id}{id}
\DeclareMathOperator{\rank}{rank}
\DeclareMathOperator{\tr}{tr}
\DeclareMathOperator{\Ad}{Ad}
\DeclareMathOperator{\vol}{vol}

\usepackage{balance}

\begin{document}

\maketitle
\thispagestyle{empty}
\pagestyle{empty}

\begin{abstract}

Differential flatness enables efficient planning and control for underactuated robotic systems, but we lack a systematic and practical means of identifying a flat output (or determining whether one exists) for an arbitrary robotic system. In this work, we leverage recent results elucidating the role of symmetry in constructing flat outputs for free-flying robotic systems. Using the tools of Riemannian geometry, Lie group theory, and differential forms, we cast the search for a globally valid, equivariant flat output as an optimization problem. An approximate transcription of this continuum formulation to a quadratic program is performed, and its solutions for two example systems achieve precise agreement with the known closed-form flat outputs. Our results point towards a systematic, automated approach to numerically identify geometric flat outputs directly from the system model, particularly useful when complexity renders pen and paper analysis intractable.
\end{abstract}

\section{Introduction}

Differential flatness is a powerful property enjoyed by many robotic systems \cite{Murray1995}, which has been employed for efficient planning and control of dynamic maneuvers \cite{Tal2021,Welde2021} including the explicit consideration of perception and actuation constraints \cite{Murali2019,Thomas2017a}. However, the construction of a flat output for complex systems such as robots is usually accomplished by human intuition or trial and error, since constructive approaches to flat output identification have traditionally been limited to special cases \cite{Rathinam1998}. 

To fill this gap, several schemes for identifying an approximate numerical flat output for a given system have been proposed. The approach of \cite{Sferrazza2016} involves generating a dataset of state and input trajectories and representing the candidate flat output and the inverse flatness diffeomorphism as weighted sums of basis functions. However, the resulting numerical flat output is only valid locally around the sampled trajectories, the basis functions chosen seem to reflect knowledge of an analytical flat output, and the approach becomes intractible for large bases. The identification of flat outputs from data in the complete absence of an analytical  system model was considered in \cite{Ma2020}, but similar locality limitations remained.

In our recent work \cite{Welde2023}, we have shown that the symmetry and Riemannian structure present in free-flying robotic systems facilitates the construction of global, equivariant, group-valued (i.e. ``geometric'') flat outputs. In particular, such flat outputs can be obtained from any section of the system's principal bundle that is orthogonal to a distribution given in terms of the kinetic energy metric and the unactuated velocities.
While this poses a geometric problem of lower dimensionality, it is not clear how to obtain closed-form solutions for such sections in general, nor are they unique, raising questions of optimality. 
In this work, we pursue the automatic identification of such geometric flat outputs, using finite element methods to solve the underdetermined partial differential equations arising from the orthogonality constraint. The problem bears similarities to the identification of ``optimal coordinates'' (i.e. an optimal trivialization) for locomotion systems \cite{Hatton2011}, from which we take inspiration.

\section{Mechanical Systems with Symmetry}

In this section, we give a brief review of essential concepts, and direct the reader to \cite{Bloch2005} for a comprehensive overview, as well as \cite{Welde2023} for an introduction to differential flatness of mechanical systems with symmetry.
We consider unconstrained mechanical systems with a Lagrangian of the form
\begin{equation}
	L : TQ \to \mathbb{R}, \, v_q \mapsto \frac{1}{2}\metric{v_q}{v_q}- P(q),
\end{equation}
subject to controlled external forces contained in $F \subseteq T^*Q$. We assume the system has ``broken'' symmetry, meaning the kinetic energy metric ${\metric{\cdot}{\cdot}}$ and the forces $F$ are respectively invariant and equivariant with respect to the free and proper action ${\Phi_g : Q \to Q}$ of a Lie group $G$, but the potential $P$ need not be. This action also induces a projection 
${\pi : Q \to S}$, where ${S = Q / G}$ is the ``shape space'' capturing the remaining ``internal'' degrees of freedom. 
The dynamics can be expressed in the form
${\nabla_{\dot{q}} \dot{q} + \mathrm{grad} \, P = {f_q}^\sharp}$, where ${f_q \in F}$
and $\nabla$ is the \textit{Riemannian} 
(or \textit{Levi-Civita}) 
\textit{connection}.

\subsection{The Mechanical Connection}

We recall that the \textit{mechanical connection} is given by \cite{Bloch2005}
\begin{equation}
	HQ = \left\{
	h_q \in TQ : \metric{h_q}{\xi_Q} = 0 \textrm{ for all } \xi \in G
	\right\},
\end{equation}
where the vector field
${	\xi_Q : Q \rightarrow TQ 
}$,
$
{
	q \mapsto 
	\left.
	\frac{d}{dt}\big(
	\Phi_{\exp t \xi} \ q 
	\big)
	\right|_{t = 0}
}$
is the \textit{infinitesimal generator} of ${\xi \in \mathfrak{g}}$.
We can also describe the mechanical connection using a $\mathfrak{g}$-valued one-form
\begin{equation}
	\mathcal{A} : TQ \to \mathfrak{g}, \, v_q \mapsto \mathcal{I}(q)^{-1} \mathcal{J}(v_q),
\end{equation}
where the \textit{momentum map} ${\mathcal{J} : TQ \to \mathfrak{g}^*}$ and the \textit{locked inertia tensor} ${\mathcal{I}(q) : \mathfrak{g} \to \mathfrak{g}^*}$ are defined by
\begin{equation}
	\pairing{\mathcal{J}(v_q)}{\xi} = \metric{v_q}{\xi_Q}, \ \ 
	\pairing{\mathcal{I}(q) \, \eta }{\xi} = \metric{\eta_Q}{\xi_Q},
\end{equation}
where $\pairing{\cdot}{\cdot}$ denotes the natural pairing of covectors and vectors.
It can be verified that $\ker \mathcal{A} = HQ$ and 
\begin{equation}
	\pairing{\mathcal{I}(g \cdot q) \, \xi}{\eta} =
	\pairing{\mathcal{I}(q) \Ad_{g^{-1}} \xi}{\Ad_{g^{-1}}\eta}.
	\label{locked_inertia_equivariance}
\end{equation}
In any trivialization $q = (s,g)$, any connection can be written
\begin{equation}
	\mathcal{A} = \Ad_g \big(
	g^{-1} dg + \mathbb{A}(s) \, ds
	\big),
	\label{local_connection_equation}
\end{equation}
where ${\mathbb{A} : TS \to \mathfrak{g}}$ is the \textit{local form} of the connection, and the \textit{local locked inertia tensor} is 
${\mathbb{I}(s) = \mathcal{I}(s,e) : \mathfrak{g} \to \mathfrak{g}^*}.$

\subsection{Trivializations, Sections, and Geometric Flat Outputs}

Any map ${y : Q \to G}$ which is equivariant i.e. $y \circ \Phi_g(q) = g \, y(q)$ induces
a \textit{trivialization}, i.e. a diffeomorphism
\begin{equation}
	\psi : Q \to S \times G, \, q \mapsto \big(\pi(q),{y}(q)\big).
\end{equation}
A \textit{section} of the bundle is any map belonging to the set
\begin{equation}
	\Gamma(Q) = \big\{
	\sigma : S \to Q : \pi \circ \sigma = \id, \, \sigma \textrm{ is smooth}
	\big\}.
\end{equation}
Relative to a fixed trivialization, any section can be given as
\begin{equation}
	\sigma : S \to S \times G, \, s \mapsto \big(s,\phi(s)\big),
	\label{phase_offset_function}
\end{equation}
using a \textit{transition map} ${\phi : S \to G}$. 
It is well-known that sections and trivializations are in bijective correspondence \cite[Prop 9.24]{Gallier2020}, and the group component of the trivialization induced by $\sigma$ is given (relative to the fixed trivialization) by
\begin{equation}
	{y} : S \times G \to G, \, (s,g) \mapsto g \, \phi(s).
	\label{group_component_section}
\end{equation}
A \textit{geometric flat output} is an equivariant map ${y : Q \to G}$ that is also a flat output of the system \cite{Welde2023}, i.e. for a smooth curve in the flat output space $G$, there is a locally unique dynamically feasible lift in the configuration manifold.\footnote{A geometric flat output may also be induced by a local (e.g. almost global) section; for brevity, here we consider trivial bundles and global geometric flat outputs, but direct the reader to \cite{Welde2023} for a broader discussion.}  

Let the \textit{unactuated subbundle} be given by ${UQ = \mathrm{coann} \, F}$. Similar to definitions in \cite{Rathinam1998} and \cite{Sato2012}, the \textit{underactuation distribution} can then be defined (as it was in \cite{Welde2023}) by
\begin{equation}
	\Delta = \mathrm{span} \big\{X, \nabla_Y X : X \in \Gamma(UQ), \ Y \in \Gamma(TQ) \big\},
\end{equation}
which is equivariant for any system with broken symmetry.
Moreover, when ${\dim G = \rank F}$ and mild assumptions hold, any solution to the constraint satisfaction problem
\begin{subequations}
	\begin{alignat}{3}
			& \textrm{find} & {\vphantom{A}\displaystyle \sigma \in \Gamma(Q)}   \ \ \ 
			\ 
		\label{find_sigma}
					\\
		&	\textrm{s.}\, \textrm{t.} \quad  
			&
			\ \ \Delta \perp \sigma(S) \subset Q
		\label{orthogonality_image}
		\end{alignat}
\end{subequations}
induces a trivialization whose group component \eqref{group_component_section} is a geometric flat output \cite{Welde2023}. Above, \eqref{orthogonality_image} enforces that $\Delta$ is everywhere orthogonal (with respect to the metric) to the image of $\sigma$ (as a submanifold of $Q$).
However, a systematic means of obtaining such a section (if it exists) in closed form is not known in the general case.

\section{A Geometric Optimization Problem}

In this section, we formulate the search for a geometric flat output in terms of the following optimization problem:
\begin{subequations}
	\begin{alignat}{3}
		\min_{\ \sigma \, \in \, \Gamma(Q)}  
		&& 
		 \ \ \ J(\sigma) \label{abstract_cost}
		&& \  (\textrm{\textit{momentum perturbation cost}})
		\\
		\textrm{s.}\, \textrm{t.} \quad  
		&& 
		 C(\sigma)
		& = 0&  (\textrm{\textit{orthogonality to }} \Delta)
		\label{constraint_functional_orthogonality}
	\end{alignat}
\end{subequations}
where the cost and constraint functionals ${J, C : \Gamma(Q) \to \mathbb{R}}$ will be constructed below, and the constraint functional will simply enforce 
\eqref{orthogonality_image}. Since in general we may have ${\rank \Delta < \dim G}$, solutions to \eqref{find_sigma}-\eqref{orthogonality_image} may not be unique, even up to the group action. We therefore propose an additional cost, which penalizes the deviation of the final geometric flat output from the ``minimum perturbation coordinates'' of the system \cite{Travers2013}, a trivialization which approximates e.g. an ``angular center of mass'' when ${G = SO(3)}$ \cite{Chen2022}. This cost will thus encourage the selection of a geometric flat output whose variation most completely captures the entire system's combined momentum, convenient for planning.

\subsection{Formulating the Cost Functional}

We first define an $\mathbb{R}$-valued bilinear form on $Q$ given by
\begin{equation}
	\rho_\mathcal{A} : 
	(u_q,v_q) \mapsto \pairing{\mathcal{I}(q) \, \mathcal{A}(v_q)}{\mathcal{A}(v_q)},
\end{equation}
which is easily seen to be symmetric and positive semidefinite. Examining the pullback of $\rho_\mathcal{A}$ by $\sigma$ given by 
\begin{equation}
	\sigma^\star \rho_\mathcal{A} : (u_s,v_s) \mapsto \pairing{\mathcal{I}(q) \, \mathcal{A} \circ d\sigma_s(u_s)}{\mathcal{A} \circ d\sigma_s(v_s)},
\end{equation}
we see that this is also a positive semidefinite symmetric bilinear form, but this time on $S$. Moreover, it will be identically zero over $S$ if and only if ${HQ = \ker \mathcal{A}}$ is tangent to $\sigma$, which can occur only if the mechanical connection is \textit{flat}
\cite{Saccon2017} and can be completely integrated to yield a trivialization that amounts to ``coordinates'' with no ``perturbation'' whatsoever.

Recall that the trace of a symmetric bilinear form $\omega$ on $S$
is given by\footnote{For all index notation, we follow the Einstein summation convention.} ${\big(\hspace{-1.7pt}\tr \omega\big)(s) = \omega({v_s}^k,{v_s}^k)}$ for any orthonormal\footnote{A $G$-invariant metric on $Q$ induces a natural metric on ${S = Q/G}$ given by
	$
	(u_s,v_s) \mapsto 
	{ \metric{(u_s)_q^{HQ}}{(v_s)_q^{HQ}}}$ for any ${q \in \pi^{-1}(s)}$,
	where ${(\cdot)^{HQ}_q : T_sS \to T_qQ}$ is the horizontal lift via the mechanical connection.
} basis ${T_sS = \spn \{{v_s}^1,\ldots,{v_s}^n\}}$, and thus the trace of a positive semidefinite form is a non-negative function which vanishes exactly at those points where the form itself vanishes. Hence, $\omega$ will vanish identically over $S$ if and only if the integral of $\tr \omega$ over $S$ vanishes. 

Therefore, we define the cost functional
\begin{equation}
	J : \Gamma(Q) \to \mathbb{R}, \, \sigma \mapsto  \int_S \tr(\sigma^\star \rho_\mathcal{A}) \, \vol_S,
	\label{cost_functional}
\end{equation}
where $\vol_S$ is the Riemannian volume form \cite[Prop. 6.4]{Gallier2020}.
The previous integrand can also be given more explicitly by
\begin{align}
	\tr (\sigma^\star \rho_\mathcal{A})(s) \nonumber
	&= \pairing{\mathbb{I}(s) \,
		\big( \phi(s)^{-1} \, d\phi_s + \mathbb{A}(s)\big)({v_s}^k) \nonumber
	}{\\& \quad \quad \quad \quad \quad 
		\big( \phi(s)^{-1} \, d\phi_s + \mathbb{A}(s)\big)({v_s}^k)
	},
\label{explicit_cost_integrand}
\end{align}
which follows from \eqref{phase_offset_function} as well as \eqref{local_connection_equation} relating the connection form $\mathcal{A}$ and the local form $\mathbb{A}$  and \eqref{locked_inertia_equivariance} describing the symmetry properties of the locked inertia tensor.

Finally, note that  although we consider minimum perturbation coordinates for the mechanical connection (for the reasons already stated), the foregoing analysis is valid for any other principal connection, including the \textit{canonical flat connection} ($\mathbb{A} = 0$) induced by some other trivialization. Such a choice would penalize (in a differential sense) the deviation of any geometric flat output obtained relative to e.g. some particular body-fixed frame.

\subsection{Reformulating the Orthogonality Constraint}

In the original constraint satisfaction problem, the orthogonality constraint \eqref{orthogonality_image} must be enforced pointwise over all of ${\sigma(S) \subset Q}$. The problem will be much more amenable to practical solution if we express this constraint using a constraint functional that vanishes exactly when the constraint holds.
To this end, we define a bilinear form
\begin{equation}
	\rho_\Delta
	:
	(u_q,v_q) \mapsto 
	\metric{\mathrm{proj}_\Delta(u_q)}{\mathrm{proj}_\Delta(v_q)},
\end{equation}
where ${\proj_\Delta : TQ \to TQ}$ is the orthogonal projection onto $\Delta$.
$\rho_\Delta$ is clearly symmetric and positive semidefinite, and 
\begin{equation}
	\Delta \perp \sigma(S)
	 \iff \sigma^\star \rho_\Delta = 0.
\end{equation}
Thus, following a line of reasoning similar to that employed in formulating the cost functional, we enforce the orthogonality constraint using the constraint functional
\begin{equation}
	C : \Gamma(Q) \to \mathbb{R}, \, \sigma \mapsto \int_S \tr(\sigma^\star \rho_\Delta) \ \vol_S,
	\label{constraint_functional}
\end{equation}
where \eqref{orthogonality_image} holds if and only if $C(\sigma) = 0$ i.e. \eqref{constraint_functional_orthogonality} holds.

To make the computation of the integrand more explicit, 
let $\Delta = \spn \{\Delta_1,\ldots,\Delta_k\}$ locally where 
$\metric{\Delta_i}{\Delta_j} = \delta_{ij}$ and each
$\Delta_i$ is equivariant. We then have
\begin{equation}
	\proj_\Delta(v_q) = \metric{\Delta_i(q)}{v_q} \, \Delta_i(q),
\end{equation}
and moreover
\begin{align}
	\rho_\Delta(u_q,v_q) 
	&= 
	\metric{u_q}{\Delta_i(q)} \, 
	\metric{\Delta_i(q)}{v_q}.
\end{align}
Since $\Delta_i$ is equivariant, there exist maps ${\eta_i : S \to \mathfrak{g}}$ and vector fields ${X_i : S \to TS}$
so that in a trivialization, we have
\begin{equation}
	\Delta_i : (s,g) \mapsto \big(X_i(s), g \, \eta_i(s)\big).
\end{equation}
Using this symmetry, we may also express the metric as 
\cite{Murray1997}
\begin{align}
	\metric{&(u_s,u_g)}{(v_s,v_g)} \nonumber \\
	& = \begin{bmatrix}
		u_s \\ g^{-1} \, u_g
	\end{bmatrix}^\textrm{T}
\setlength\arraycolsep{1pt}
	\begin{bmatrix}
		m(s) & \mathbb{A}(s)^\textrm{T} \mathbb{I}(s)
	\\
		\mathbb{I}(s) \mathbb{A}(s) & \mathbb{I}(s) 
	\end{bmatrix}
	\begin{bmatrix}
		 v_s \\ 
		g^{-1} \, v_g
	\end{bmatrix},
\end{align}
where ${u_q = (u_s,u_g)},$ ${v_q = (v_s,v_g)} \in TS \times TG$.
With this fact in mind, a straightforward computation yields
\begin{align}
	&\tr (\sigma^\star \rho_\Delta)(s)  \nonumber \\
	&
	=  \hspace{-1pt}
		\begin{bmatrix}
			{v_s}^k
			\\
			\phi(s)^{-1} \, d\phi_s ({v_s}^k) 
		\end{bmatrix}
		^\textrm{T}
		\hspace{-7pt}
		\mathbb{P}_i(s)^\textrm{T}
		\mathbb{P}_i(s)
		\hspace{-2.5pt}
		\begin{bmatrix}
			{v_s}^k
			\\
			\phi(s)^{-1} \, d\phi_s ({v_s}^k)
		\end{bmatrix}\hspace{-2.5pt},
	\label{explicit_constraint_integrand}
\end{align}
where we define 
\begin{align}
	\mathbb{P}_i(s) &=
	\begin{bmatrix}
		X_i(s) \\
		\eta_i(s) 
	\end{bmatrix}^\textrm{T}
	\setlength\arraycolsep{1pt}
	\begin{bmatrix}
	m(s) & \mathbb{A}(s)^\textrm{T} \mathbb{I}(s)
	\\
	\mathbb{I}(s) \mathbb{A}(s) & \mathbb{I}(s) 
\end{bmatrix}
	.
\end{align}

\subsection{Gauge Freedom}

To study the symmetry of the proposed cost and constraint, consider an inherited $G$-action on $\Gamma(Q)$ given by 
\begin{equation}
	\Psi_g : \Gamma(Q) \to \Gamma(Q), \, \sigma \mapsto \big(s \mapsto \Phi_g \circ \sigma(s)\big).
\end{equation}
In a trivialization, this action is described by
\begin{equation}
	\Psi_g \circ \Big(s \mapsto \big(s,\phi(s)\big)\Big) \ = \ s \mapsto \big(s,g \, \phi(s)\big),
\end{equation}
and moreover we may compute 
\begin{align}
	\big(g \, \phi(s)\big)^{-1} \, d (g \, \phi)_s(v_s) 
	&= \phi(s)^{-1} g^{-1} \, g \, d\phi_s(v_s)  \\ &=  \phi(s)^{-1} \, d\phi_s(v_s).
\end{align}
Thus, in view of the fact that the only dependence of \eqref{explicit_cost_integrand} and \eqref{explicit_constraint_integrand} on $\phi$ (and therefore, $\sigma$) takes the previous form, $J$ and $C$ are invariant under $\Psi_g$.
Therefore, if $\sigma$ is a solution to the problem \eqref{abstract_cost}-\eqref{orthogonality_image}, so is $\Psi_g(\sigma)$ for all $g \in G$.
Thus, either the addition of a symmetry-breaking cost or constraint can resolve this ambiguity, or this gauge freedom can be freely adjusted after solving the problem once, without re-solving.

\section{Finite Element Methods}

We now transcribe our abstract optimization problem to a finite dimension. Inspired by \cite{Hatton2011}, we parameterize the candidate section \eqref{phase_offset_function} using a transition map of the form
\begin{equation}
	\phi : s \mapsto \exp w_i \zeta_i(s),
	\label{phase_offset_parametrization}
\end{equation}
where each ${w_i \in \mathbb{R}}$ is a weight to be optimized and ${\zeta_i : S \to \mathfrak{g}}$ is a basis function corresponding to a multivariate polynomial spline over a hypercube mesh (see \cite{Hatton2011} for a related formulation with piecewise linear elements and a simplex mesh). Basis functions with high-order continuity are desirable, since e.g. two derivatives are necessary to express the group component of the system's acceleration.\footnote{Note that Hermite polynomial basis functions with $C^2$ smoothness have been given up to at least three dimensions \cite{Boateng2023}, but may be available in higher dimensions as well. Since for our purposes, $\dim S = \rank UQ$, we are concerned with low-dimensional cases, an advantage of approximating the section $\sigma: S \to Q$ instead of the flat output ${y}: Q \to G$ directly.
}  

\subsection{First Order Approximation of the Exponential Map}

For simplicity, we make the approximation $\exp \xi \approx e + \xi$ for small $\xi$ (which is exact when $G$ is Abelian),
so that
\begin{align}
	\phi(s) &\approx e +  w_i \zeta_i(s), \\
	\phi(s)^{-1}  d\phi_s(v_s) & \approx \big(e - w_i \zeta_i(s)\big)\big(w_i d\zeta_i(v_s)\big) \\
	&
	\approx w_i d\zeta_i(v_s). 
\end{align}
The cost functional integrand \eqref{explicit_cost_integrand} is thus approximated by
\begin{align}
		\tr &(\sigma^\star \rho_\mathcal{A})(s) 
		\nonumber
		\\ 
	&\approx \pairing{\mathbb{I}(s) \, 
		\big( w_i d\zeta_i + \mathbb{A}\big)({v_s}^k)
	}{
		\big( w_j d\zeta_j + \mathbb{A}\big)({v_s}^k)
	},
	\label{approximate_cost_integrand}
\end{align}
and similarly for the constraint functional integrand \eqref{explicit_constraint_integrand},
\begin{align}
	\tr& (\sigma^\star \rho_\Delta)(s)  \nonumber \\
	&\approx 
	\begin{bmatrix}
		{v_s}^k \\
		w_h d\zeta_h({v_s}^k)
	\end{bmatrix}
	^\textrm{T}
	\mathbb{P}_i(s)^\textrm{T}
	\mathbb{P}_i(s)
	\begin{bmatrix}
		{v_s}^k \\
		w_j d\zeta_j({v_s}^k)
	\end{bmatrix}.
\label{approximate_constraint_integrand}
\end{align}

\subsection{Gaussian Quadrature Approximation of the Integrals}

We employ Gaussian quadrature \cite{Jayan2014} to approximate the integrals in the functionals. Considering coordinates $s^i$ for $S = {\bigcup_{j=1}^m E_j}$ where $E_j$ are finitely many disjoint elements, 
\begin{align}
	\int_S f(s) \vol_S 
	&= \sum_{j=1}^m \int_{E_j} \hspace{-2pt} \Big(
	\hspace{-1pt}
		f(s) 
		\sqrt{\vphantom{\big|}\det \mathbb{G}(s)}
	\hspace{1pt}
 \Big) \, ds^1 \hspace{-3pt} \ldots ds^n \\ 
	&\approx \sum_{j=1}^m \sum_{i=1}^p c_{ij} \, \Big(
	f(s_{ij})
		\sqrt{\det \mathbb{G}(s_{ij})} \,
	 \Big),
\end{align}
where each ${c_{ij} \in \mathbb{R}}$ and ${s_{ij} \in S}$ is a predetermined optimal weight and sampling point respectively, and $\mathbb{G}(s)$ is the induced Riemannian metric in the coordinates $s^i$. 
Since Gaussian quadrature with $p$ samples is exact when the integrand
 is a polynomial of degree $2p-1$ or less,
we expect this to be a good approximation if our mesh is sufficiently fine and $\zeta_i$ is a polynomial of degree $p$ or less in $s^i$.

\subsection{Approximate Transcription to a Quadratic Program}

It is clear that pointwise over $S$, both \eqref{approximate_cost_integrand} and \eqref{approximate_constraint_integrand} are quadratic forms in the weights $w_i$. Since quadrature approximates integrals via a weighted sum, our final approximations of \eqref{cost_functional} and \eqref{constraint_functional} can be expressed in the form
\begin{align}
	J(\sigma) &\approx \tfrac{1}{2} w_i A_{ij} w_j + {b_i} w_i + c, 
	\label{approx_cost_quadratic_form}
	\\
	C(\sigma) &\approx 
	\tfrac{1}{2} w_i D_{ij} w_j + {e_i} w_i + f = 0,
	 \label{approx_constraint_quadratic_form}
\end{align}
where $A_{ij}$ and $D_{ij}$ are positive semidefinite. Due to transcription, \eqref{approx_constraint_quadratic_form} may be infeasible even if a geometric flat output exists. However, because \eqref{approx_constraint_quadratic_form} is nonnegative in view of \eqref{approximate_constraint_integrand}, we relax the orthogonality constraint to simply enforce minimization of $C(\sigma)$ instead, where minimization to zero (when feasible) corresponds to exact constraint satisfaction.

Thus, we obtain an approximate transcription of the continuum problem \eqref{abstract_cost}-\eqref{constraint_functional_orthogonality} to a Quadratic Program (QP):
\vspace{-10pt}
\begin{subequations}
	\begin{alignat}{4}
		\min_{w \, \in \, \mathbb{R}^N}  &&  
		\	\ \ \tfrac{1}{2} w_i A_{ij} w_j + {b_i}&  w_i + c   && \ \ (\textit{momentum cost}) \label{qp_cost_functional}\\ 
		\textrm{s.}\,\textrm{t.} &&  	D_{ij} w_j + e_i = & \, 0 && \ \ (\textit{orthogonality}) \\ 
		&& 
		w_i \zeta_i(s_0)  = & \log(g_0)
		&&\ \ (\textit{breaks symmetry})
		\label{qp_gauge_freedom}
	\end{alignat}
\end{subequations}
where \eqref{qp_gauge_freedom} is added to resolve the gauge freedom.
The QP is sparse due to the choice of basis functions with local support, permitting efficient solution even for very fine meshes.

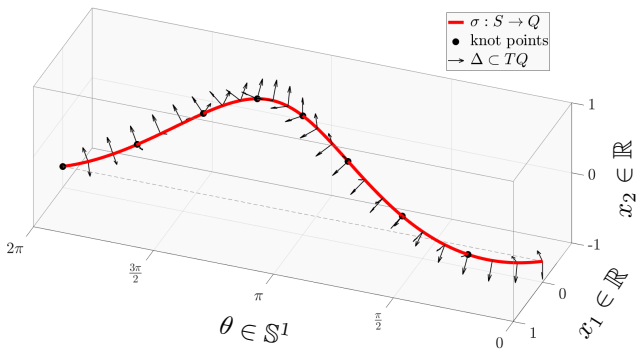
\begin{figure}
\begin{tikzpicture}
	\node (img) {
		\includegraphics[clip,trim={5.5cm 1cm 0cm 1.5cm},width=\columnwidth]{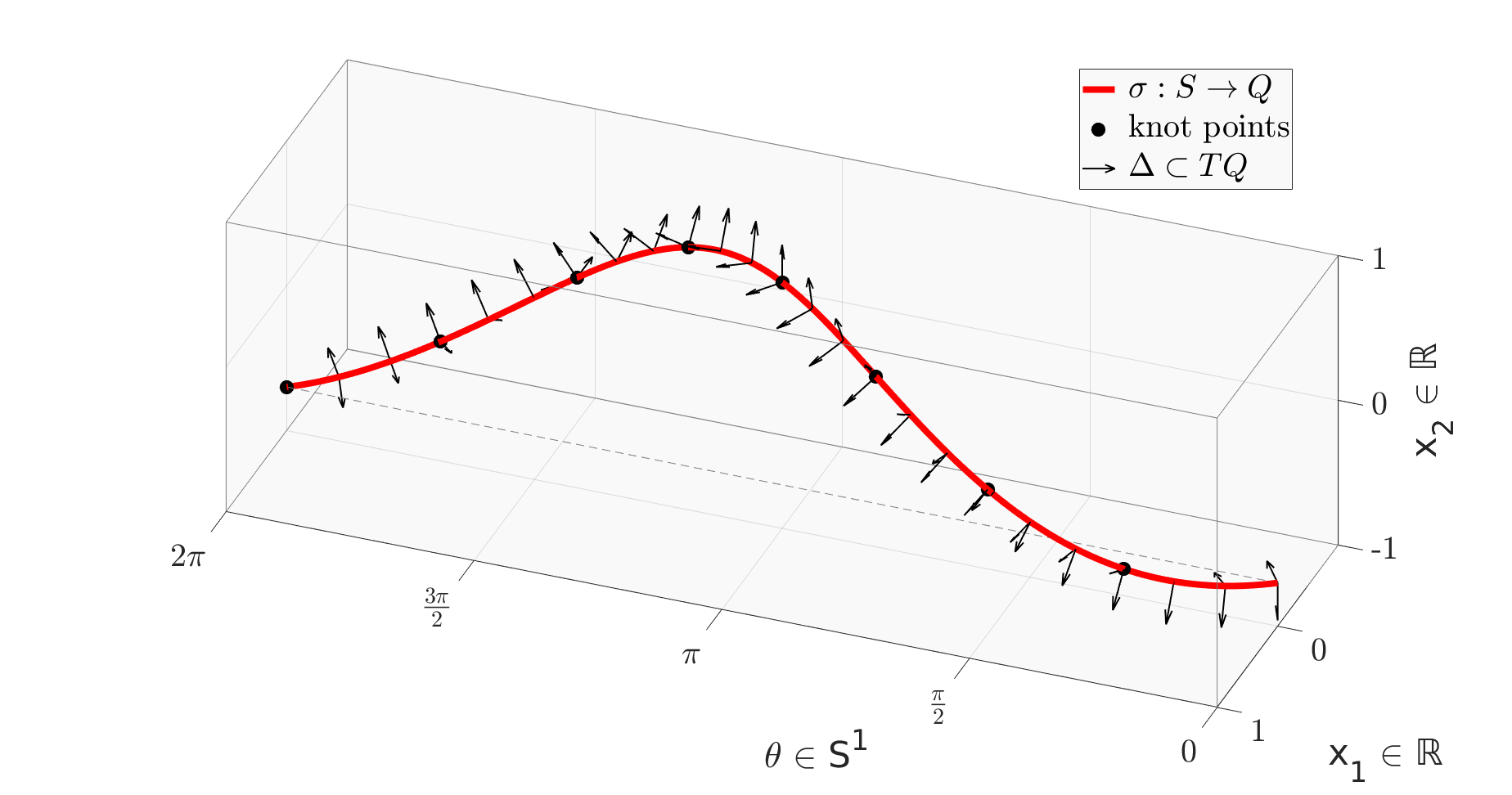}	
	};
	\node [fill=white] at (-.4,-2.3){$\phantom{x_1 \in \mathbb{R}}$};
	\node [rotate=-10.5] at (-1.,-2.05){$\theta \in \mathbb{S}^1$};
	\node [fill=white] at (3.6,-2.1
	){$\phantom{x_1 \in \mathbb{R}}$};
	\node [rotate=58] at (3.6,-1.7){${ x_1 \in \mathbb{R}}$};
	\node [rotate=90,fill=white] at (3.9,-.05){$x_2 \in \mathbb{R}$};
\end{tikzpicture}
\vspace{-14pt}
\caption{
	\label{rocket_section}
	For the Planar Rocket \cite[Ex. 1]{Welde2023}, we depict the finite element boundaries (i.e. the knot points of the quintic spline) and
	the restriction of $\Delta$ to ${\sigma(\mathbb{S}^1) \subset SE(2)}$, which can be seen to be orthogonal to 
	the solution ${\sigma : \mathbb{S}^1 \to SE(2)}$  to \eqref{qp_cost_functional}-\eqref{qp_gauge_freedom}. (Technically, we plot ${\Delta^\flat \subset T^* Q}$, and vectors and covectors appearing at right angles amounts to orthogonality.)
}
\vspace{-16pt}
\end{figure}
\begin{figure}
	\begin{minipage}{.4\textwidth}
		\centering
		\includegraphics{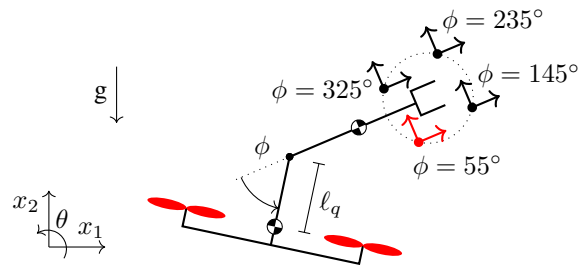}
	\end{minipage}%
	\vspace{-12pt}
	\caption{
		\label{aerial_manipulator_output}
		Visualization of the numerical geometric flat output obtained for the Planar Aerial Manipulator\cite[Ex. 2]{Welde2023}, which amounts to a frame parallel to the end effector with a $\phi$-dependent offset. When $\ell_q =0$, the offset goes to zero. Physical parameters have been chosen to exaggerate the offset for clarity. The frame in red corresponds to the angle $\phi$ visualized above.
	}
	\vspace{-16pt}
\end{figure}

\section{Results}

We implement the proposed method to identify geometric flat outputs on two benchmark systems, namely the Planar Rocket \cite[Ex. 1]{Welde2023} on $SE(2)$ (an $\mathbb{R}^2$ bundle over $\mathbb{S}^1$) and the Planar Aerial Manipulator \cite[Ex. 2]{Welde2023} on $SE(2) \times \mathbb{S}^1$ (an $SE(2)$ bundle over $\mathbb{S}^1$). 
In both examples, we employ $C^2$ basis functions corresponding to quintic Hermite splines over an evenly divided mesh on $\mathbb{S}^1$ with ${m=8}$ elements, and perform Gaussian quadrature with ${p=6}$ sample points per element. The orthogonal section obtained is visualized in Fig. \ref{rocket_section} for the Planar Rocket, meanwhile the final geometric flat output is visualized in Fig. \ref{aerial_manipulator_output} for the Planar Aerial Manipulator. With suitable (and easily chosen) gauge freedom constraints, the numerical geometric flat outputs identified for both systems show precise numerical agreement with the closed form geometric flat outputs previously given in \cite{Welde2023}.

\section{Conclusion}

In this work, we have used Riemannian geometry and Lie group theory to cast the search for a geometric flat output as an optimization problem. Explicit cost and constraint functionals are formulated, which are shown to be naturally invariant with respect to the system's symmetry. We perform an approximate transcription of the continuum problem to a finite dimensional quadratic program using finite element methods, Gaussian quadrature, and a small angle approximation. The approach is implemented for two example systems, and close agreement with closed-form solutions provides evidence in favor of the method's feasibility for accurate identification of geometric flat outputs for more complex systems, for which closed form solutions are unknown.

In our ongoing work, we will pursue an iterative approach, where at each iteration, the exponential map is linearized about the current guess until convergence. Since this process can be performed offline, the computational requirements should not be an obstacle.
In order to apply the approach to more complex systems, we will explore computing the relevant quantities (e.g. the underactuation distribution $\Delta$ and the mechanical connection $\mathcal{A}$) numerically from a standard modeling format (e.g. a URDF \cite{Willow2009}), adapting efficient multibody dynamics solvers (e.g. Pinocchio \cite{Pinocchio}) to perform the necessary covariant differentations by using a suitable factorization of the Coriolis forces \cite{Echeandia2021}. 
We will also explore the extent to which the relaxed orthogonality constraint that simply minimizes $C(\sigma)$ may yield an ``approximate flat output'' when a truly orthogonal section does not exist, even in the continuum setting.
Finally, our approach amounts to a continuous deformation of an existing section; thus, a potential pitfall is that in general, all sections of a principal bundle need not be homotopic, although over a contractible shape space (e.g. a local region), all sections are homotopic.

\balance 
\bibliographystyle{IEEEtran}
\bibliography{IEEEabrv,refs}

\begin{thebibliography}{10}
\providecommand{\url}[1]{#1}
\csname url@rmstyle\endcsname
\providecommand{\newblock}{\relax}
\providecommand{\bibinfo}[2]{#2}
\providecommand\BIBentrySTDinterwordspacing{\spaceskip=0pt\relax}
\providecommand\BIBentryALTinterwordstretchfactor{4}
\providecommand\BIBentryALTinterwordspacing{\spaceskip=\fontdimen2\font plus
\BIBentryALTinterwordstretchfactor\fontdimen3\font minus
  \fontdimen4\font\relax}
\providecommand\BIBforeignlanguage[2]{{%
\expandafter\ifx\csname l@#1\endcsname\relax
\typeout{** WARNING: IEEEtran.bst: No hyphenation pattern has been}%
\typeout{** loaded for the language `#1'. Using the pattern for}%
\typeout{** the default language instead.}%
\else
\language=\csname l@#1\endcsname
\fi
#2}}

\bibitem{Murray1995}
R.~M. Murray, M.~Rathinam, and W.~Sluis, ``Differential flatness of mechanical
  control systems: A catalog of prototype systems,'' \emph{ASME International
  Mechanical Engineering Congress and Exposition}, 1995.

\bibitem{Tal2021}
E.~Tal and S.~Karaman, ``Accurate tracking of aggressive quadrotor trajectories
  using incremental nonlinear dynamic inversion and differential flatness,''
  \emph{IEEE Transactions on Control Systems Technology}, vol.~29, no.~3, pp.
  1203--1218, 2021.

\bibitem{Welde2021}
J.~Welde, J.~Paulos, and V.~Kumar, ``Dynamically feasible task space planning
  for underactuated aerial manipulators,'' \emph{IEEE Robotics and Automation
  Letters}, vol.~6, pp. 3232--3239, 2021.

\bibitem{Murali2019}
V.~Murali, I.~Spasojevic, W.~Guerra, and S.~Karaman, ``Perception-aware
  trajectory generation for aggressive quadrotor flight using differential
  flatness,'' in \emph{2019 American Control Conference (ACC)}, 2019, pp.
  3936--3943.

\bibitem{Thomas2017a}
J.~Thomas, J.~Welde, G.~Loianno, K.~Daniilidis, and V.~Kumar, ``Autonomous
  flight for detection, localization, and tracking of moving targets with a
  small quadrotor,'' \emph{IEEE Robotics and Automation Letters}, vol.~2, pp.
  1762--1769, 2017.

\bibitem{Rathinam1998}
M.~Rathinam and R.~M. Murray, ``Configuration flatness of {L}agrangian systems
  underactuated by one control,'' \emph{SIAM Journal on Control and
  Optimization}, vol.~36, pp. 164--179, 1998.

\bibitem{Sferrazza2016}
C.~Sferrazza, D.~Pardo, and J.~Buchli, ``Numerical search for local (partial)
  differential flatness,'' \emph{IEEE International Conference on Intelligent
  Robots and Systems}, vol. 2016-Novem, pp. 3640--3646, 2016.

\bibitem{Ma2020}
S.~F. Ma, G.~Leylaz, and J.-Q. Sun, ``Identification of differentially flat
  output of underactuated dynamic systems,'' \emph{International Journal of
  Control}, vol.~0, pp. 1--27, 2020.

\bibitem{Welde2023}
J.~Welde, M.~D. Kvalheim, and V.~Kumar, ``The role of symmetry in constructing
  geometric flat outputs for free-flying robotic systems,'' in \emph{IEEE
  International Conference on Robotics and Automation (ICRA)}, 2023.

\bibitem{Hatton2011}
R.~L. Hatton and H.~Choset, ``Geometric motion planning: The local connection,
  {S}tokes' theorem, and the importance of coordinate choice,''
  \emph{International Journal of Robotics Research}, vol.~30, pp. 988--1014, 7
  2011.

\bibitem{Bloch2005}
A.~M. Bloch, \emph{Nonholonomic Mechanics and Control}, 1st~ed.\hskip 1em plus
  0.5em minus 0.4em\relax Springer, 2015, vol.~24.

\bibitem{Gallier2020}
J.~Gallier and J.~Quaintance, \emph{Differential Geometry and Lie Groups, A
  Second Course, no. 13 in Geometry and Computing}.\hskip 1em plus 0.5em minus
  0.4em\relax Springer, 2020.

\bibitem{Sato2012}
K.~Sato and T.~Iwai, ``Configuration flatness of {L}agrangian control systems
  with fewer controls than degrees of freedom,'' \emph{Systems and Control
  Letters}, vol.~61, pp. 334--342, 2012.

\bibitem{Travers2013}
M.~Travers, R.~Hatton, and H.~Choset, ``Minimum perturbation coordinates on
  {SO(3)},'' in \emph{2013 American Control Conference}, 2013, pp. 2006--2012.

\bibitem{Chen2022}
Y.-M. Chen, G.~Nelson, R.~Griffin, M.~Posa, and J.~Pratt, ``Angular center of
  mass for humanoid robots,'' \emph{arXiv preprint arXiv:2210.08111}, 2022.

\bibitem{Saccon2017}
A.~Saccon, S.~Traversaro, F.~Nori, and H.~Nijmeijer, ``On centroidal dynamics
  and integrability of average angular velocity,'' \emph{IEEE Robotics and
  Automation Letters}, vol.~2, pp. 943--950, 2017.

\bibitem{Murray1997}
R.~M. Murray, ``Nonlinear control of mechanical systems: a {L}agrangian
  perspective,'' \emph{Annual Reviews in Control}, vol.~21, pp. 31--42, 1997.

\bibitem{Boateng2023}
H.~A. Boateng and K.~Bradach, ``Triquintic interpolation in three dimensions,''
  \emph{Journal of Computational and Applied Mathematics}, p. 115254, 2023.

\bibitem{Jayan2014}
S.~Jayan and K.~Nagaraja, ``Numerical integration over n-dimensional cubes
  using generalized gaussian quadrature,'' in \emph{Proc. Jangjeon Math. Soc},
  vol.~17, 2014, pp. 63--69.

\bibitem{Willow2009}
{Willow Garage, Inc.}, ``{Unified Robot Description Format},''
  http://wiki.ros.org/urdf/, 2009.

\bibitem{Pinocchio}
J.~Carpentier, F.~Valenza, N.~Mansard, \emph{et~al.}, ``Pinocchio: fast forward
  and inverse dynamics for poly-articulated systems,''
  https://stack-of-tasks.github.io/pinocchio, 2015--2023.

\bibitem{Echeandia2021}
S.~Echeandia and P.~M. Wensing, ``{Numerical Methods to Compute the Coriolis
  Matrix and Christoffel Symbols for Rigid-Body Systems},'' \emph{Journal of
  Computational and Nonlinear Dynamics}, vol.~16, no.~9, 07 2021.

\end{thebibliography}

\end{document}